\documentclass[runningheads]{llncs}
\usepackage[T1]{fontenc}
\usepackage{graphicx}
\usepackage{caption}
\usepackage{subcaption}
\usepackage{amsmath}
\usepackage{amsfonts}
\usepackage{tabularx}
\usepackage{afterpage}
\usepackage{hyperref}

\usepackage{color}

\begin{document}
\title{Structure-preserving Image Translation for Depth Estimation in Colonoscopy}

\author{Shuxian Wang\orcidID{0000-0002-0184-3021} \and
Akshay Paruchuri\orcidID{0000-0003-4664-3186}
\and
Zhaoxi Zhang\orcidID{0009-0007-7193-1426}
\and
Sarah McGill\orcidID{0000-0002-4006-2703}\and
Roni Sengupta\orcidID{0000-0001-5914-3469}}
% index{Wang, Shuxian}
% index{Paruchuri, Akshay}
% index{Zhang, Zhaoxi}
% index{McGill, Sarah}
% index{Sengupta, Roni}
\authorrunning{S. Wang et al.}
\institute{University of North Carolina at Chapel Hill, Chapel Hill NC 27514, USA}

\maketitle              % typeset the header of the contribution
\begin{abstract}
Monocular depth estimation in colonoscopy video aims to overcome the unusual lighting properties of the colonoscopic environment. One of the major challenges in this area is the domain gap between annotated but unrealistic synthetic data and unannotated but realistic clinical data. Previous attempts to bridge this domain gap directly target the depth estimation task itself. We propose a general pipeline of structure-preserving synthetic-to-real (sim2real) image translation (producing a modified version of the input image) to retain depth geometry through the translation process.
This allows us to generate large quantities of realistic-looking synthetic images for supervised depth estimation with improved generalization to the clinical domain. We also propose a dataset of hand-picked sequences from clinical colonoscopies to improve the image translation process. We demonstrate the simultaneous realism of the translated images and  preservation of depth maps via the performance of downstream depth estimation on various datasets.

\end{abstract}
\section{Introduction}
Colorectal cancer (CRC) is one of the leading causes of cancer mortality in the United States; the American Cancer Society estimates that there will be over 150,000 new cases and 50,000 deaths in 2024. Increased screening is one of the factors contributing to reductions in mortality \cite{siegel-cancer-2024}. Optical colonoscopy is the gold standard method for CRC screening but its effectiveness is highly dependent upon the skill of the physician performing the examination \cite{nierengarten-colonoscopy-2023}. Around 20\% of potentially pre-cancerous polyps are missed during colonoscopies \cite{vanRijn-colonoscopy-2006}\cite{vemulapalli-polyps-2022}.

3D reconstruction from optical colonoscopy video can improve efficacy via guidance and visualization to the physician, automatic measurements, and autonomous navigation.
One of the major challenges in this area is the lack of realistic data suitable for training neural networks to perform depth and pose estimation. While synthetic \cite{simcol3d} and phantom \cite{c3vd} datasets exist, they do not accurately represent the reflectance properties of \textit{in vivo} tissue. Previous approaches towards closing the domain gap \cite{mahmood_deep_2018}\cite{rau-implicit-2019}\cite{wang-surface-2023} do not target challenging viewpoints making up the majority of colonoscopy videos. In this work, we propose an image translation method that generates realistic-looking video frames from synthetic colonoscopies while preserving the depth information and without requiring complex modeling of mucus and \textit{in vivo} tissue. In this way, we are able to bridge the gap between unrealistic synthetic data with dense ground truth depth annotation and realistic but un-annotated clinical data to improve depth estimation on unseen clinical data. In addition, we introduce two new datasets of manually selected frames from clinical colonoscopies representing viewpoints that are particularly challenging for depth estimation and downstream reconstruction. This data both improves the realism of our image translation results and provides a dataset against which to test the quality of depth estimation results.
Code is available at \url{github.com/sherry97/struct-preserving-cyclegan}
and data at \url{endoscopography.web.unc.edu}

\section{Related Work}
\begin{figure}[b]
    \centering
    \begin{subfigure}[t]{0.44\textwidth}
        \centering
        \includegraphics[angle=180,origin=c,width=0.3\textwidth]{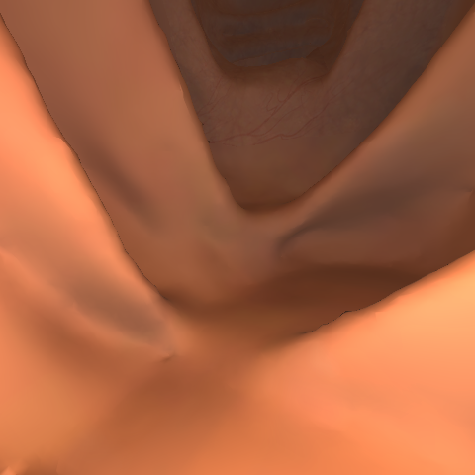}
        \includegraphics[trim={135 0 135 0},clip,width=0.3\textwidth]{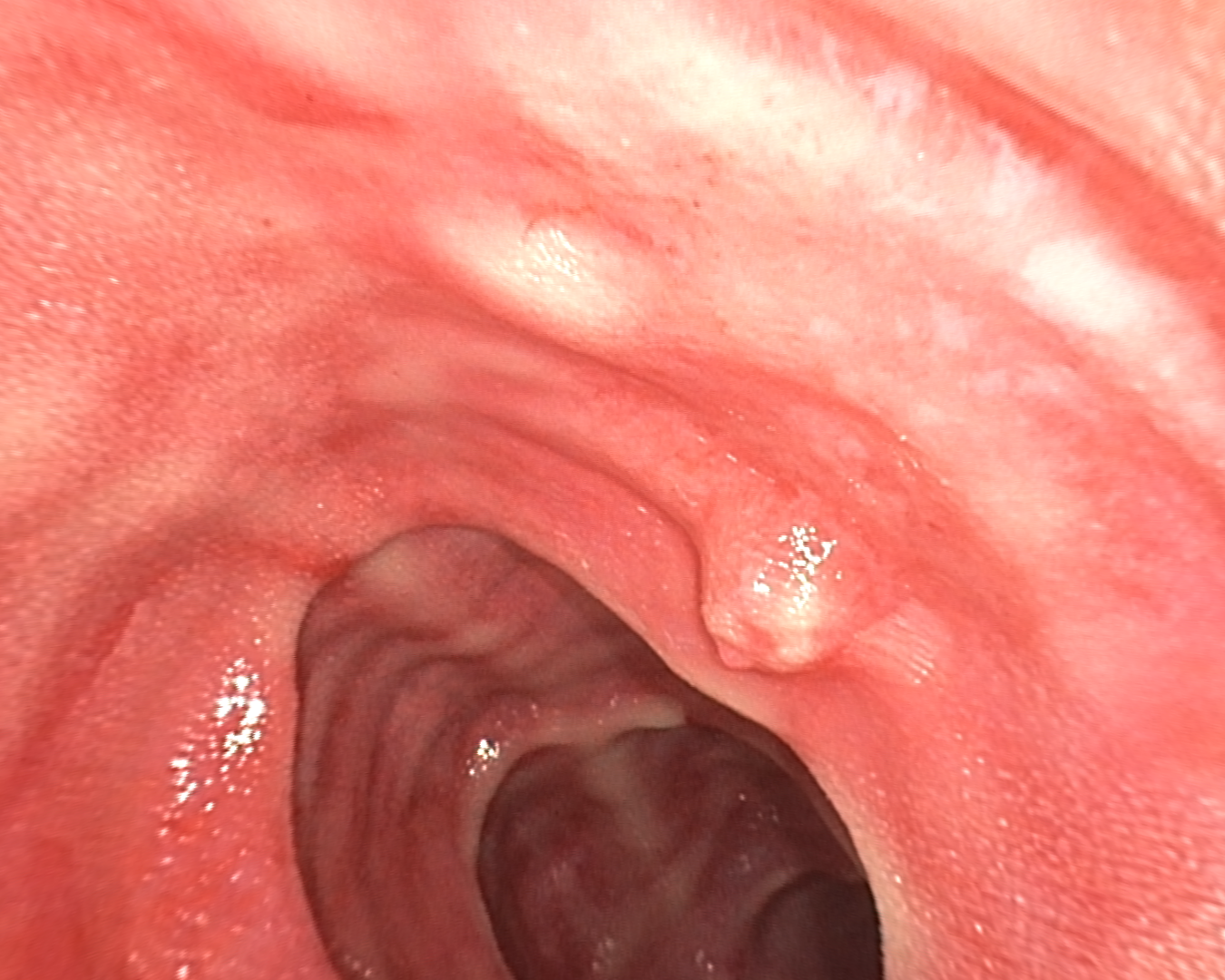}
        \includegraphics[angle=270,origin=c,trim={27 0 27 0},clip,width=0.3\textwidth]{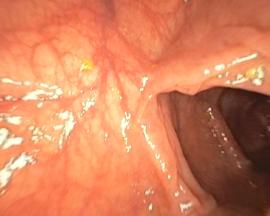}
        \caption{Textures from SimCol3D \cite{simcol3d} (left), C3VD \cite{c3vd} (center), and proposed oblique dataset (right).}
    \end{subfigure}
    \hfill
    \begin{subfigure}[t]{0.54\textwidth}
        \centering
        \includegraphics[width=0.3\textwidth]{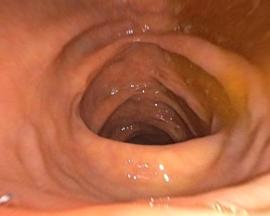}
        \includegraphics[width=0.3\textwidth]{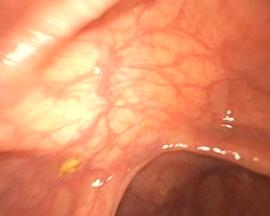}
        \includegraphics[width=0.3\textwidth]{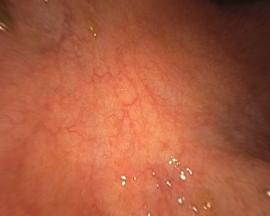}
        \caption{Viewpoint categories in colonoscopy: axial (left, Colon10K \cite{colon10k}), oblique (center), and \textit{en face} (right).}
        \label{fig:viewpoints}
    \end{subfigure}
    \caption{Sample frames from various datasets.}
    \label{fig:data_demo}
\end{figure}

Prior datasets targeting reconstruction from colonoscopy come from clinical procedures (EndoMapper \cite{endomapper}, Colon10K \cite{colon10k}), fully synthetic procedures (SimCol3D \cite{simcol3d}, Zhang et al. \cite{zhang-sydneysimulator-2021}), or robotic colonoscopy of a silicone phantom model of the colon (C3VD \cite{c3vd}). Clinical data by nature does not have per-frame depth or pose annotations; while synthetic and phantom data have such annotations, the geometry and light reflectance properties of living tissue is challenging to replicate synthetically and therefore the textures present in the synthetic and phantom data are notably different from those observed in clinical practice (Fig. \ref{fig:data_demo}). While the use of image translation to bridge the synthetic to clinical domain gap has been addressed previously (Sec. \ref{sec:background_domgap}), we propose a general modular framework particularly targeting depth estimation (Sec. \ref{sec:background_depth}) on challenging viewpoints.
This is the first work that performs structure-preserving image translation from the synthetic to clinical colonoscopy domain without requiring a pre-trained depth estimator or feature extractor in the target clinical domain.

\subsection{Domain gap}\label{sec:background_domgap}
Using image translation for colonoscopic depth estimation, Rau et al. \cite{rau-implicit-2019} propose image-to-depth translation to directly estimate depths from images. In contrast, Mahmood and Durr \cite{mahmood_deep_2018} combine synthetic depth estimation with real-to-synthetic image translation at inference.

For other tasks, Chen et al. \cite{chen-beyond-2022} propose a structure-preserving image-to-image generative adversarial network (GAN) to improve segmentation using mutual information in the latent encoding. Similarly, Yoon et al. \cite{yoon_colonoscopic_2022} propose using GAN-based dataset augmentation to boost performance.

For general-purpose image translation, many previous works build upon CycleGAN \cite{CycleGAN2017} due to the structure preservation implicit in the cyclical architecture.
Cheng et al. \cite{cheng_structure-preserving_2020} present a structure-preserving alternative that decomposes style (extracted via a pretrained autoencoder) from structure (extracted via a pretrained monocular depth estimator).

\subsection{Depth estimation}\label{sec:background_depth}
In order to demonstrate the effectiveness of our image translation approach, we use performance on monocular depth estimation as the metric for comparison. Wang et al. \cite{wang-surface-2023} propose a self-supervised extension of Monodepth2 \cite{monodepth2} for the colonoscopy domain with an iterative refinement step. For general depth estimation, modern Transformer-based methods \cite{eftekhar2021omnidata}\cite{depthanything} demonstrate high-quality depth estimation results on non-medical data but rely on large training datasets.

\section{Data}
Generally, we can categorize the viewpoint of a single frame as axial, oblique, or \textit{en face} (Fig. \ref{fig:viewpoints}). Oblique and \textit{en face} viewpoints can be challenging for depth estimation due to the lack of strong geometric features. However, they make up about 70\% of non-obfuscated frames within a colonoscopy video so reliable depth estimation from these views and their subsequent incorporation into reconstruction provides significant additional information about surface geometry over reconstruction from axial views alone.

In this work, we introduce two distinct datasets: the first of oblique views and the second of \textit{en face} views. Both consist of sequences of consecutive frames manually selected from a library of video recordings of full colonoscopy procedures. The datasets have been curated on the basis of the viewpoint of each frame such that a sequence extends as long as each consecutive frame is of the same viewpoint category modulo gaps of up to 30 consecutive frames with excessive obfuscation (e.g. water drops on the lens).

All frames are pre-processed in the same manner. Using computed camera intrinsics and the Matlab undistortFisheyeImage function, we warp fisheye projection into a pinhole projection. We then crop the image to remove the unused image area and resize to $270 \times 216$ pixels. The original videos were recorded using CF and PCF series Olympus colonoscopes with a raw image size of $1350 \times 1080$.
The UNC Office of Human Research Ethics has determined that this work does not constitute human subject research and does not require Internal Review Board approval.

\paragraph{Oblique dataset}
The first dataset, which we call the oblique dataset, consists of sequences manually selected to exclude obfuscated frames, fully axial views, and fully \textit{en face} views. There are 93 sequences totalling 16,756 frames. Each sequence has between 2 and 586 frames, averaging 180 frames per sequence. We randomly divide this dataset into $90\%$ train and $10\%$ test partitions with divisions being made at the sequence (rather than frame) level.

\paragraph{\textit{En face} dataset}
The second dataset, which we call the \textit{en face} dataset, consists of sequences manually selected to exclude obfuscated frames and contain only fully \textit{en face} views. There are 14 sequences totalling 816 frames. Each sequence has been 15 and 136 frames, averaging 58 frames per sequence. In this work, we only use this dataset for evaluation due to its small size.

\section{Methods}
\begin{figure}[t]
    \centering
    \includegraphics[trim={50 260 200 25},clip,width=.8\textwidth]{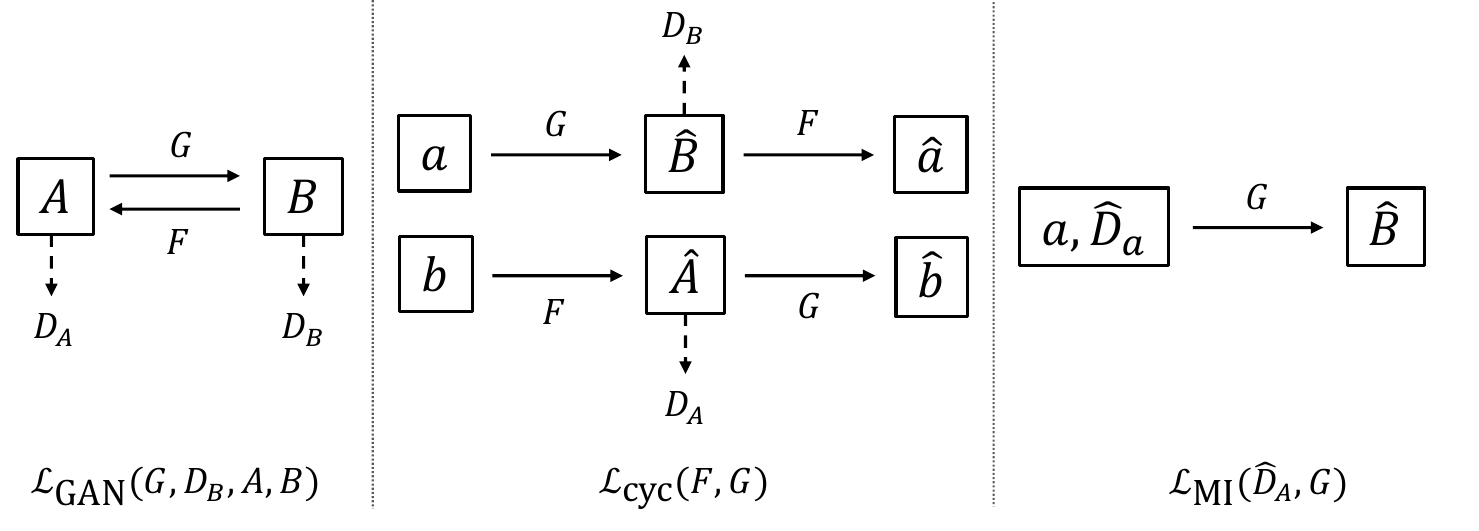}
        \caption{Image translation framework with image domains $A$ and $B$, generators $G:A \rightarrow B$ and $F: B \rightarrow A$, and discriminators $D_A$ and $D_B$. Let $a \in A, b \in B$ denote data samples and let $\hat{D}_a$ denote the depth map corresponding to sample $a$. Downstream depth estimation uses output of generator $G(A)$.}
    \label{fig:framework}
\end{figure}

We demonstrate the realism of the image translation result and effectiveness of our proposed structure-preserving loss term via downstream depth estimation. Fig. \ref{fig:framework} illustrates our framework. Our image translation result is additionally improved with the use of our proposed data over pre-existing datasets.

\paragraph{Image translation}

We use the standard CycleGAN losses with generator $G: A \rightarrow B$ and discriminator $D_B$ (and similarly generator $F: B \rightarrow A$ and discriminator $D_A$):
\begin{equation}
    \mathcal{L}_{\text{GAN}}(G, D_B, A, B) = \mathbb{E}_{b \sim p_{\text{data}}(b)}[\log D_{B}(b)] + \mathbb{E}_{a \sim p_{\text{data}}(a)}[\log(1 - D_{B}(G(a)))]
\end{equation}
\begin{equation}
    \mathcal{L}_{\text{cyc}}(G,F) = \mathbb{E}_{a \sim p_{\text{data}}(a)}[ || F(G(a)) - a ||_1] + \mathbb{E}_{b \sim p_{\text{data}}(b)}[ || G(F(b)) - b ||_1]
\end{equation}

In order to explicitly constrain the translation to preserve depth information so that the depths and translated image pairs can be used to train supervised depth estimation, we add mutual information loss. Mutual information circumvents the recursive problem of depth estimation or feature extraction on challenging clinical data.
\begin{equation}
    \mathcal{L}_{\text{MI}}(\hat{D}_A, G) = \mathbb{E}_{a \sim p_{\text{data}}(a)} \sum_{i=1}^{|\hat{D}_a|} \sum_{j=1}^{|\hat{I}(G(a))|} \frac{| \hat{D}_i \cap \hat{I}_j|}{N} \log \frac{N|\hat{D}_i \cap \hat{I}_j|}{|\hat{D}_a||\hat{I}(G(a))|}
\end{equation}

where $\hat{D}_a$ is the ground truth depth map corresponding to the input sample $a$, $\hat{I}(\cdot)$ is image intensity (average of all color channels), and $N$ is the number of total combinations.
We discretize the data into 256 bins for both depths and intensity. This loss is only applied for A $\rightarrow$ B translation.
Our full objective is:

\begin{align*}
    \mathcal{L}(G,F,D_A,D_B) = &\lambda_{\text{GAN}} \mathcal{L}_{\text{GAN}}(G,D_B,A,B) + \lambda_{\text{GAN}} \mathcal{L}_{\text{GAN}}(F, D_A, B, A) \\
    &+ \lambda_{\text{cyc}} \mathcal{L}_{\text{cyc}}(G,F) + \lambda_{\text{MI}} \mathcal{L}_{\text{MI}}(\hat{D}_A, G)
\end{align*}

We train CycleGAN to perform image translation with SimCol3D as domain A and our proposed oblique dataset as domain B. Table \ref{tab:ablations} describes the ablations in this portion used for downstream depth estimation.

\paragraph{Depth estimation}
A significant mismatch between the translated image and the original depth map (lack of structure preservation during translation) will result in poor depth estimation generalization for any model.
Here we are interested in depth estimation performance as a metric for the structure preservation through the image translation process and therefore note any architecture could be used. 
We use the Monodepth2 \cite{monodepth2} architecture trained fully supervised from scratch. We pair the RGB result from image translation with the depth map from the original synthetic data for labels.
In order to avoid data overlap, we measure performance of all models on C3VD \cite{c3vd}.
We convert the fisheye projection to a pinhole projection using the OpenCV undistort function.

\paragraph{Implementation details}
For image translation, we train the modified CycleGAN using four NVIDIA Titan Xp GPUs for 30 epochs. We use the Adam optimizer and initial learning rate of 2e-4. We use weights $\lambda_{\text{GAN}} = 10.0$, $\lambda_{\text{cyc}} = 0.5$, and $\lambda_{\text{MI}} = 1.0$.

For depth estimation with Monodepth2, we use a Resnet34 backbone and train the model using a NVIDIA Quadro RTX 5000 for 20 epochs with mean squared error loss, Adam optimizer, and initial learning rate of 1e-4. We use data augmentations of random cropping to $256 \times 256$ and random horizontal and vertical flipping. At inference, we rescale images to $256 \times 256$. All code is implemented using Pytorch.

\section{Results}
\begin{table}[t]
    \centering
    \caption{Image translation metrics against oblique dataset. Using $\mathcal{L}_\text{MI}$ helps the model produce images more similar to the  distribution of test images.}
    \begin{tabularx}{\textwidth}{|c|*2{>{\centering\arraybackslash}X|}}
        \hline
         \textbf{Model} & \textbf{Frechet Inception Distance $\downarrow$} & \textbf{Kernel Inception Distance $\downarrow$} \\
         \hline
         CycleGAN & 2.225 & $0.220 \pm 0.0179 $\\
         Ours & 0.300 & $0.090 \pm 0.0146 $\\
         \hline
    \end{tabularx}
    \label{tab:translation_metrics}
\end{table}

\begin{figure}[t]
\centering 
    \begin{tabularx}{\textwidth}{c *6{>{\centering\arraybackslash}X}}
    \rotatebox[origin=l]{90}{\begin{tabularx}{0.56\textwidth}{*4{>{\centering\arraybackslash}X}}
        &   Closest & & Input \\
        CycleGAN & Oblique & Ours & (SimCol3D) \\
    \end{tabularx}}
    &
    \begin{subfigure}{0.14\textwidth}
        \includegraphics[width=\textwidth]{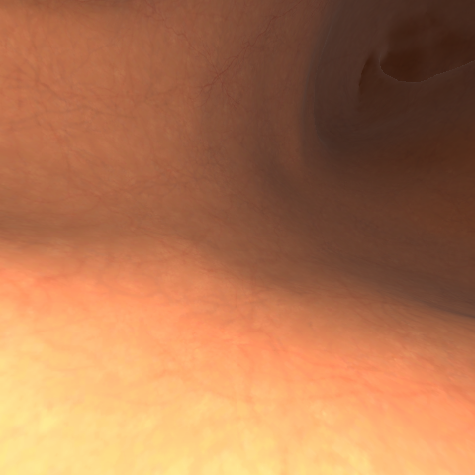}
        \includegraphics[width=\textwidth]{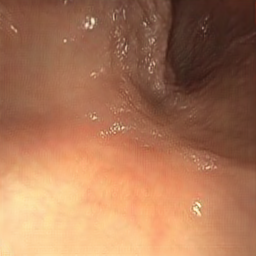}
        \includegraphics[trim={27 0 27 0},clip,width=\textwidth]{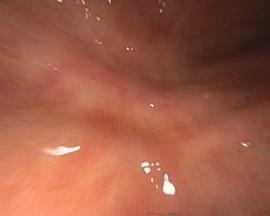}
        \includegraphics[width=\textwidth]{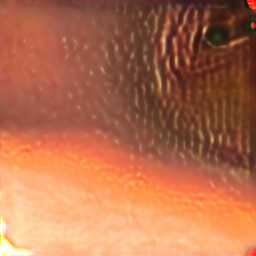}
    \end{subfigure}
    &
    \begin{subfigure}{0.14\textwidth}
        \includegraphics[width=\textwidth]{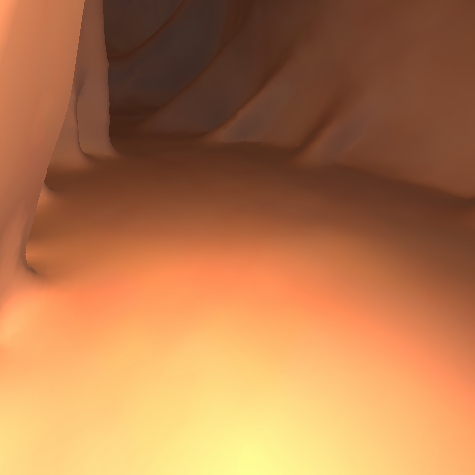}
        \includegraphics[width=\textwidth]{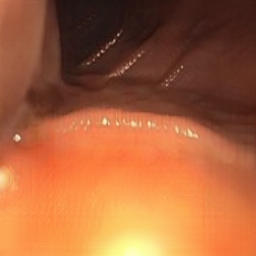}
        \includegraphics[trim={27 0 27 0},clip,width=\textwidth]{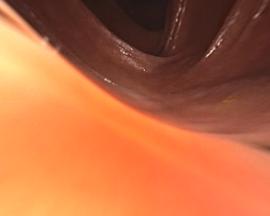}
        \includegraphics[width=\textwidth]{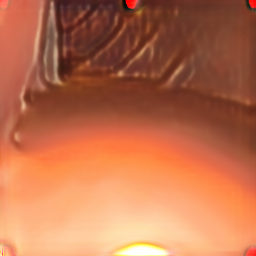}
    \end{subfigure}
    &
    \begin{subfigure}{0.14\textwidth}
        \includegraphics[width=\textwidth]{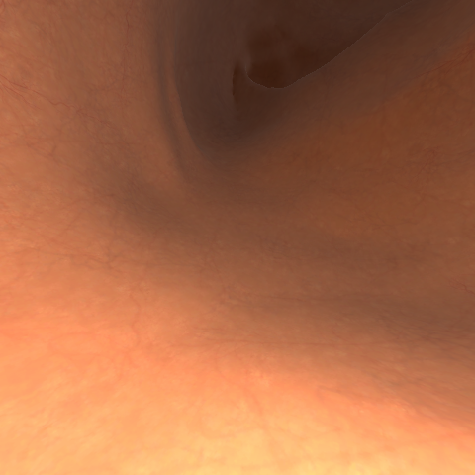}
        \includegraphics[width=\textwidth]{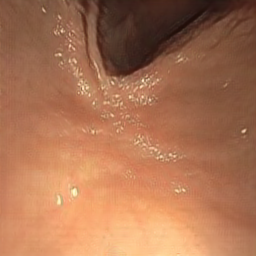}
        \includegraphics[trim={27 0 27 0},clip,width=\textwidth]{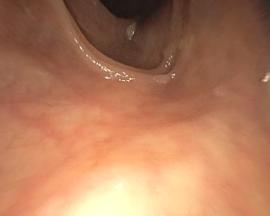}
        \includegraphics[width=\textwidth]{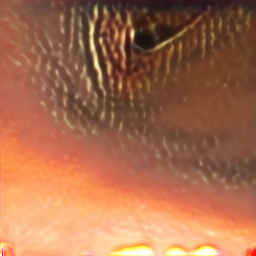}
    \end{subfigure}
    &
    \begin{subfigure}{0.14\textwidth}
        \includegraphics[width=\textwidth]{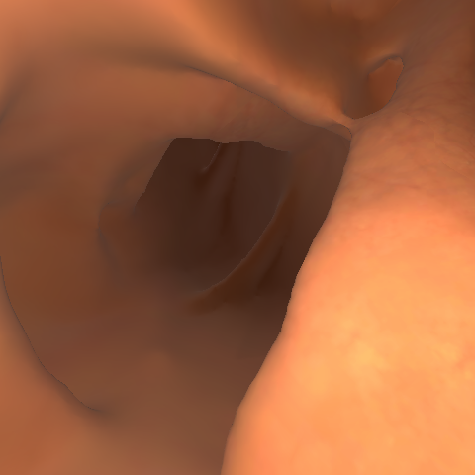}
        \includegraphics[width=\textwidth]{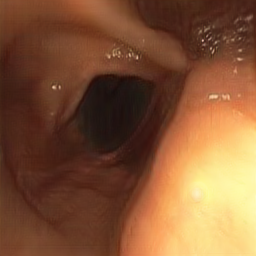}
        \includegraphics[trim={27 0 27 0},clip,width=\textwidth]{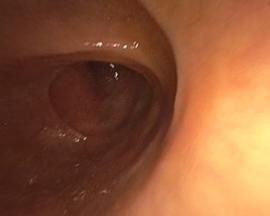}
        \includegraphics[width=\textwidth]{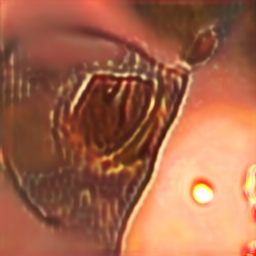}
    \end{subfigure}
    &
    \begin{subfigure}{0.14\textwidth}
        \includegraphics[width=\textwidth]{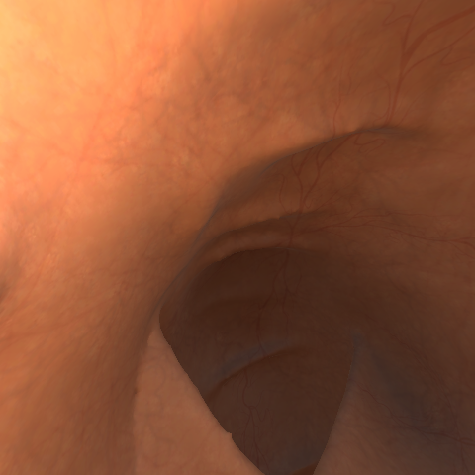}
        \includegraphics[width=\textwidth]{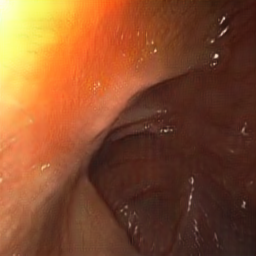}
        \includegraphics[trim={27 0 27 0},clip,width=\textwidth]{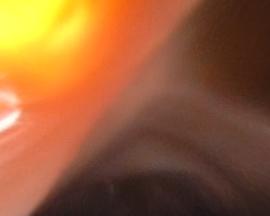}
        \includegraphics[width=\textwidth]{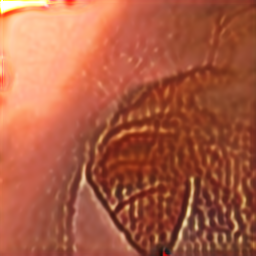}
    \end{subfigure}
    &
    \begin{subfigure}{0.14\textwidth}
        \includegraphics[width=\textwidth]{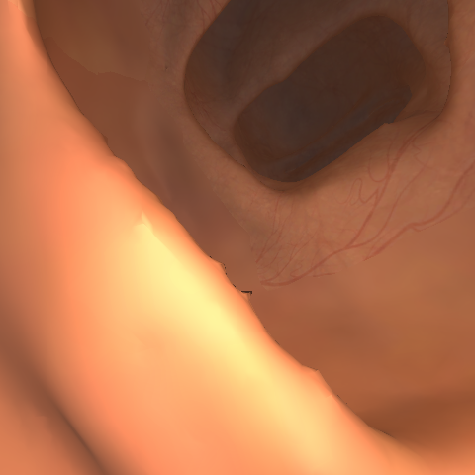}
        \includegraphics[width=\textwidth]{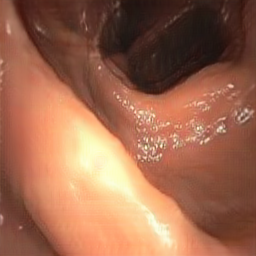}
        \includegraphics[trim={27 0 27 0},clip,width=\textwidth]{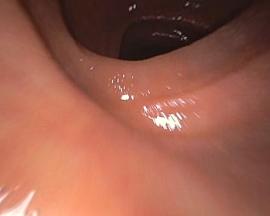}
        \includegraphics[width=\textwidth]{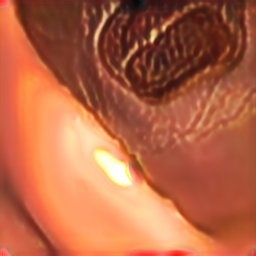}
    \end{subfigure}
    \end{tabularx}
    \caption{Examples comparing the SimCol3D input frame, our translation, closest image in oblique dataset via SSIM, and translation with vanilla CycleGAN.}
    \label{fig:translation_demo}
\end{figure}

\subsection{Image translation}
We find that the translation result (Fig. \ref{fig:translation_demo}) has both improved texture realism and retains the overall geometry of the input image. Most notably, the translation adds the specular points missing from SimCol3D without explicit representation. The specularity is distributed in a manner consistent with our expectation that surfaces closer to the camera and having surface normal directions parallel with the viewing direction will exhibit more specular effects than those either farther from the camera or with normal direction different from that of the viewing direction.
In Table \ref{tab:translation_metrics}, we compare translation metrics against translation with $\lambda_\text{MI}=0$ (vanilla CycleGAN) and find the metrics support our perception of improved translation results when $\lambda_\text{MI} > 0$.

\subsection{Depth estimation}

We measure depth estimation performance on C3VD \cite{c3vd} for comparison against baseline models due to its better realism compared to other options but note that the textures and geometries represented in that dataset remain different from those observed in clinical practice. Our qualitative assessment of depth predictions on our proposed oblique and full \textit{en face} datasets demonstrate a notable performance gap on realistic images.

\paragraph{C3VD} In Table \ref{tab:c3vd}, we provide metrics computed after median rescaling to adjust depth scale across models. For zero-shot models (relying on generalization), we find that our framework produces the best performance in most metrics. We also find that the performance is similar across various models and training datasets. We conclude that the performance on this dataset is satisfactory given the architecture and simplicity of the evaluation dataset, and look for a larger performance gap on more challenging clinical frames.

\begin{table}[t]
    \centering
    \caption{Ablations on translation target dataset and use of MI loss. All depth estimations use Monodepth2 architecture, varying in input data.} \label{tab:ablations}
    \begin{tabularx}{\textwidth}{|*3{>{\centering\arraybackslash}X|}}
        \hline
        \textbf{Depth Estimation Input} & \textbf{Translation Domain B} & \textbf{Uses MI Loss} \\
        \hline
        Baseline (no translation) & - & - \\
        Ours & oblique & \checkmark \\
        \hline
        Ours$_\text{CG}$ & oblique & - \\
        Ours$_\text{C10K}$ & Colon10K & \checkmark \\
        Ours$_\text{C3VD}$ & C3VD & \checkmark \\
        \hline
    \end{tabularx}
    
    \caption{Depth evaluation on C3VD (mm). Best categorical performance highlighted. Multi-shot models train on C3VD while zero-shot rely on generalization. On easy data like C3VD, all experiments perform similarly.}
    \label{tab:c3vd}
    \begin{tabularx}{\textwidth}{|X|c|c|c|c|c|c|}
        \hline
        \centering \textbf{Category} & \textbf{Model} & \textbf{RMSE} $\downarrow$ & \textbf{Abs}$_{\text{rel}} \downarrow$ & $\delta < 1.25 \uparrow$ & $\delta < 1.25^2 \uparrow$ & $\delta < 1.25^3 \uparrow$ \\ 
        \hline
        & Monodepth2 \cite{eftekhar2021omnidata} & 18.640 & 0.297 & 0.490 & 0.731 & 0.861 \\ 
        \centering Multi-shot & UNet \cite{eftekhar2021omnidata} & \textbf{5.520} & \textbf{0.090} & \textbf{0.917} & \textbf{0.994} & \textbf{0.999} \\ 
        & Ours$_\text{C3VD}$ & 7.250 & 0.150 & 0.794 & 0.968 & 0.996 \\ 
        \hline
        & NormDepth \cite{wang-surface-2023} & 7.401 & \textbf{0.169} & 0.731 & 0.948 & 0.997 \\ 
        & Baseline & 9.847 & 0.205 & 0.626 & 0.934 & 0.991 \\ 
        \centering Zero-shot & Ours$_\text{C10K}$ & 8.089 & 0.174 & 0.735 & 0.958 & 0.995 \\ 
        & Ours$_\text{CG}$ & 7.636 & 0.174 & 0.730 & \textbf{0.960} & \textbf{0.998} \\ 
        & Ours & \textbf{7.209} & 0.174 & \textbf{0.738} & 0.948 & 0.994 \\
        \hline
    \end{tabularx}
\end{table}

\begin{figure}[t]
    \centering
    {\begin{tabularx}{0.85\textwidth}{*6{>{\centering\arraybackslash}X}}
         Input & Ours & Ours$_\text{C10K}$ & Ours$_\text{C3VD}$ & Baseline & NormDepth
    \end{tabularx}}
    \includegraphics[width=0.85\textwidth]{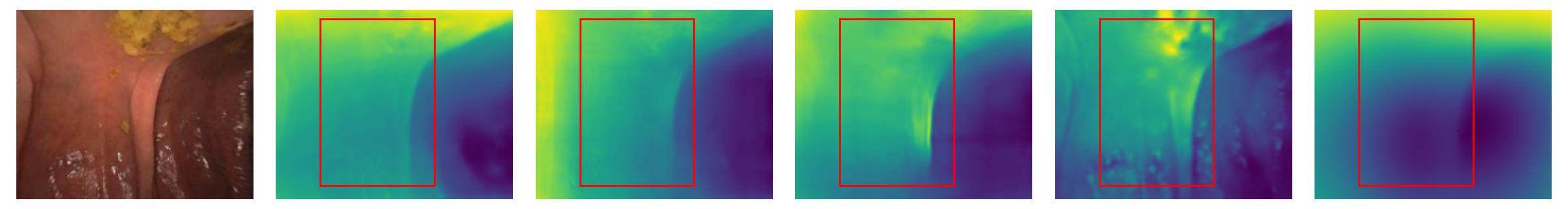}
    \includegraphics[width=0.85\textwidth]{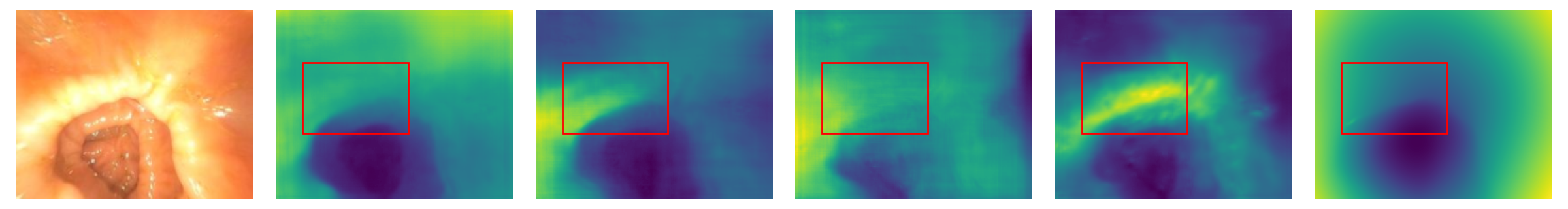}
    \includegraphics[width=0.85\textwidth]{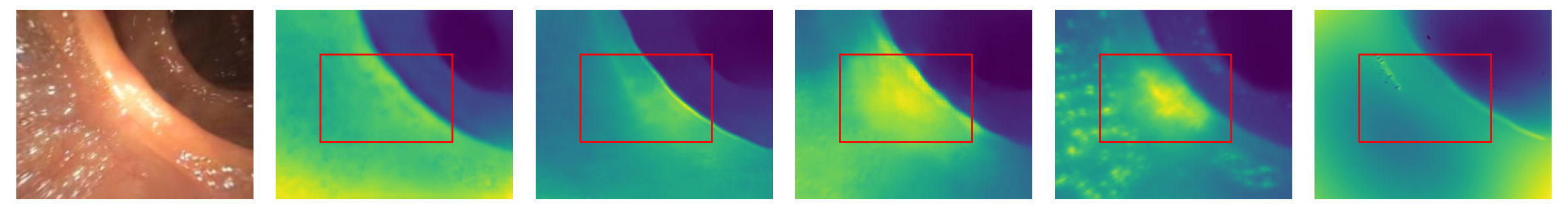}
    \caption{Depth estimation on oblique dataset. Boxes highlight differences. Image translation framework improves monocular depth estimation in general, with best performance using our proposed dataset as translation target.}
    \label{fig:depth_demo_monodepth}

    {\begin{tabularx}{0.85\textwidth}{*6{>{\centering\arraybackslash}X}}
         Input & Ours & Ours$_\text{C10K}$ & Ours$_\text{C3VD}$ & Baseline & NormDepth
    \end{tabularx}}
    \includegraphics[width=0.85\textwidth]{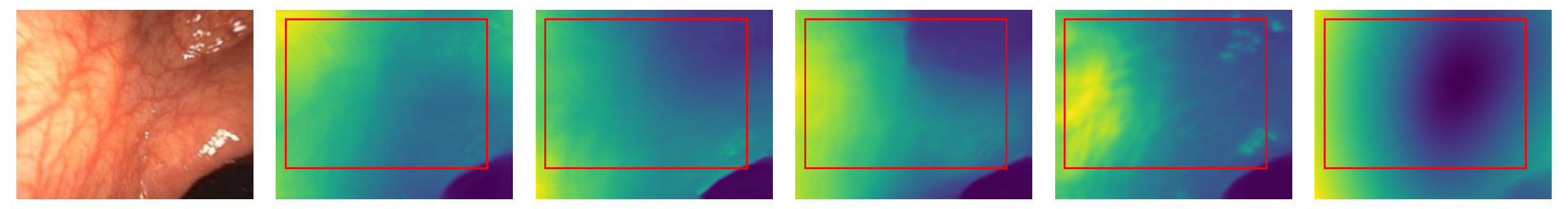}
    \includegraphics[width=0.85\textwidth]{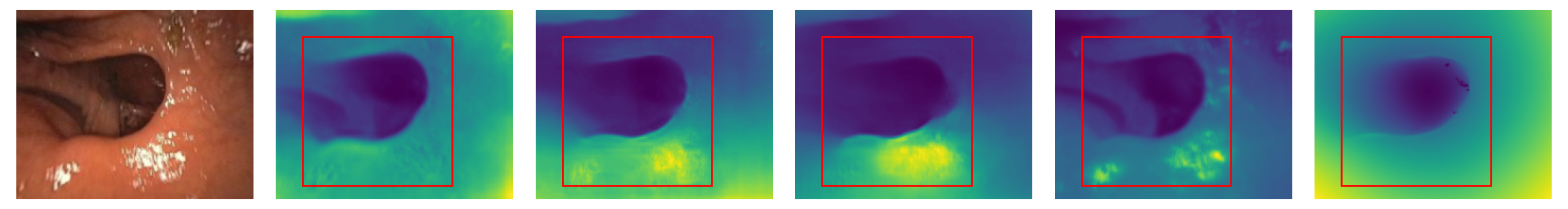}
    \includegraphics[width=0.85\textwidth]{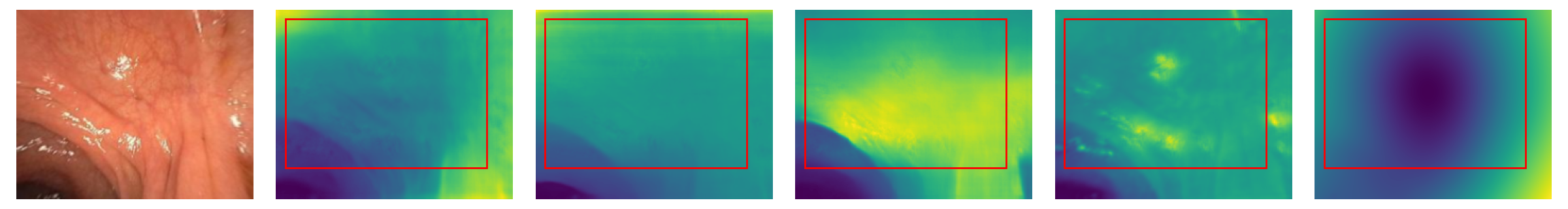}
        \caption{Depth estimation on \textit{en face} dataset. Boxes highlight differences. Notable improvements from our framework on frames with few geometric features.}
    \label{fig:enface_depth_demo_monodepth}
\end{figure}

\begin{figure}
    \centering
    {\begin{tabularx}{0.8\textwidth}{*6{>{\centering\arraybackslash}X}}
         Input & Ours & Ours$_\text{C10K}$ & Ours$_\text{C3VD}$ & Baseline & NormDepth
    \end{tabularx}}
    \includegraphics[width=0.8\textwidth]{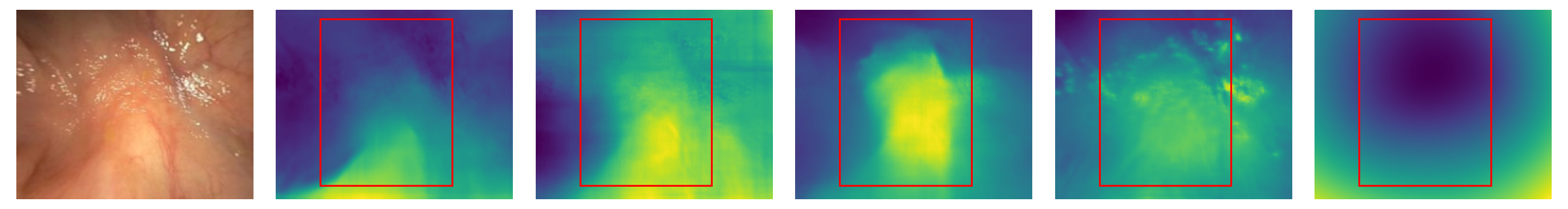}
    \includegraphics[width=0.8\textwidth]{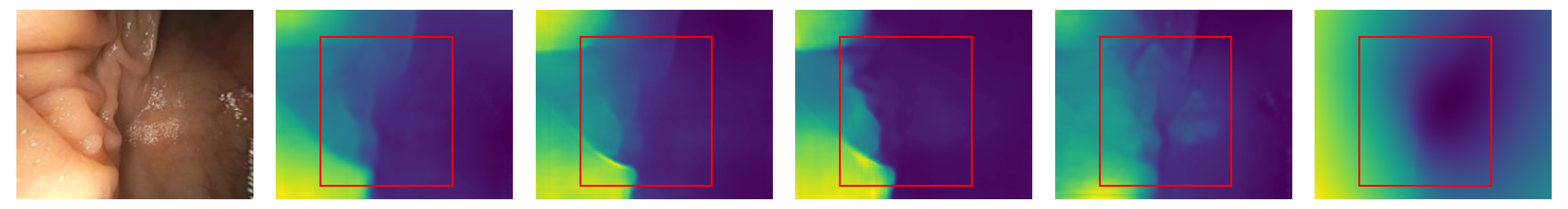}
    \includegraphics[width=0.8\textwidth]{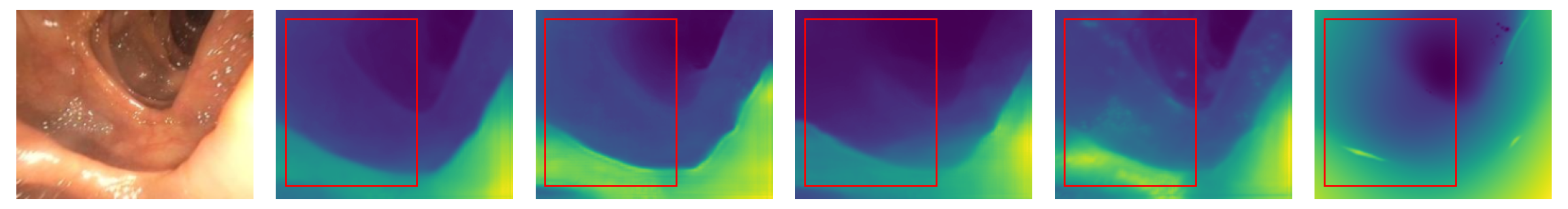}
    \includegraphics[width=0.8\textwidth]{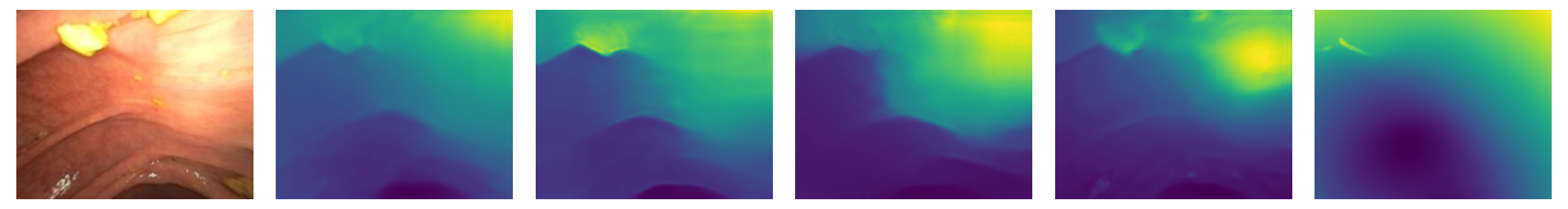}
    \includegraphics[width=0.8\textwidth]{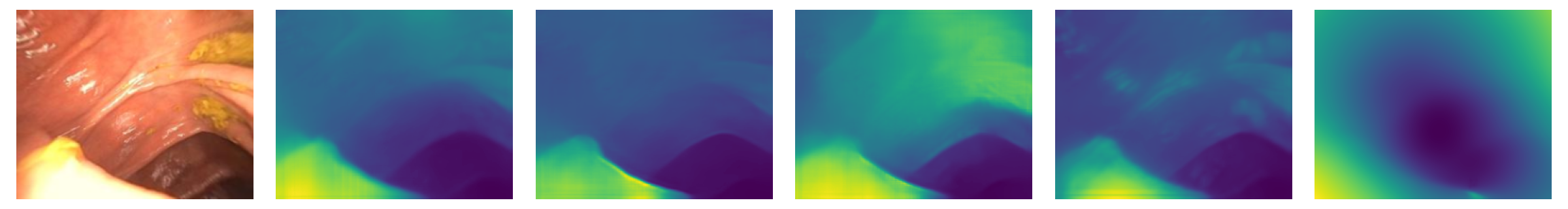}
    \includegraphics[width=0.8\textwidth]{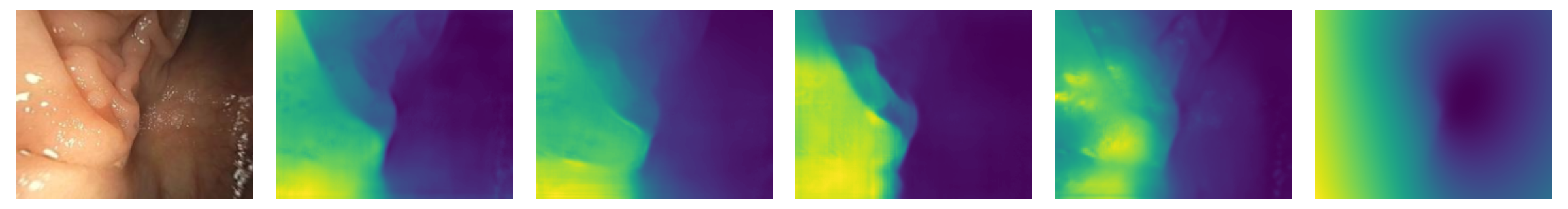}
    \includegraphics[width=0.8\textwidth]{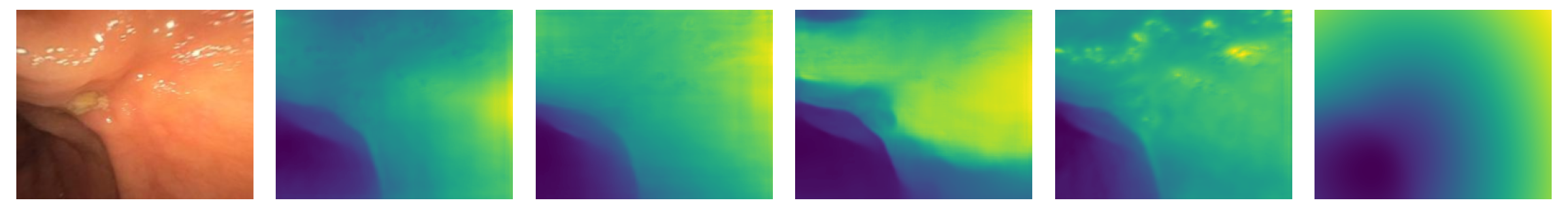}
    \includegraphics[width=0.8\textwidth]{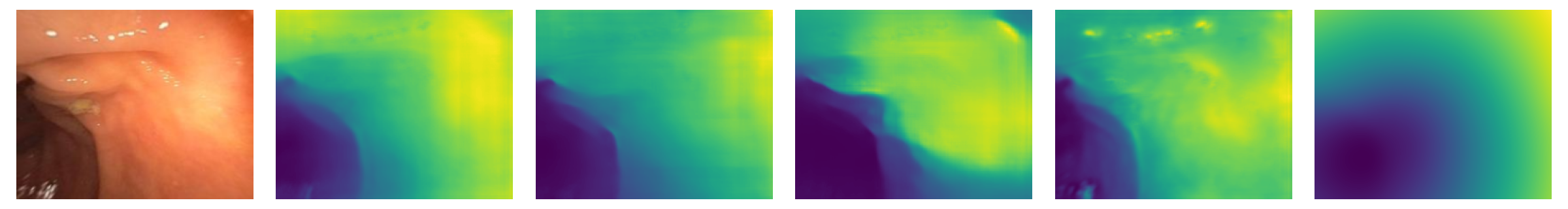}
    \includegraphics[width=0.8\textwidth]{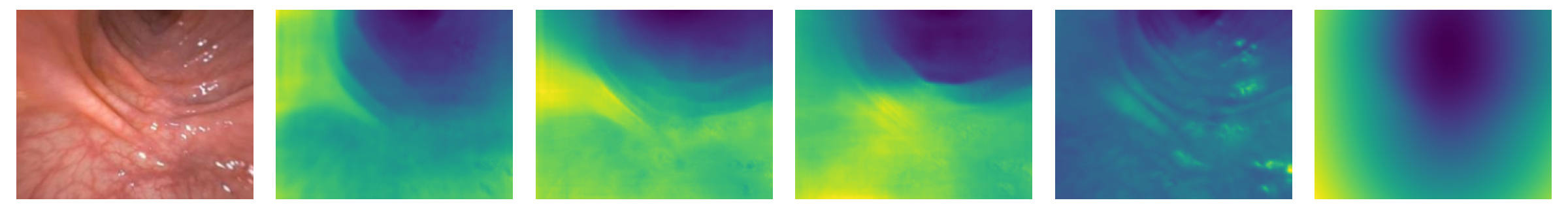}
    \includegraphics[width=0.8\textwidth]{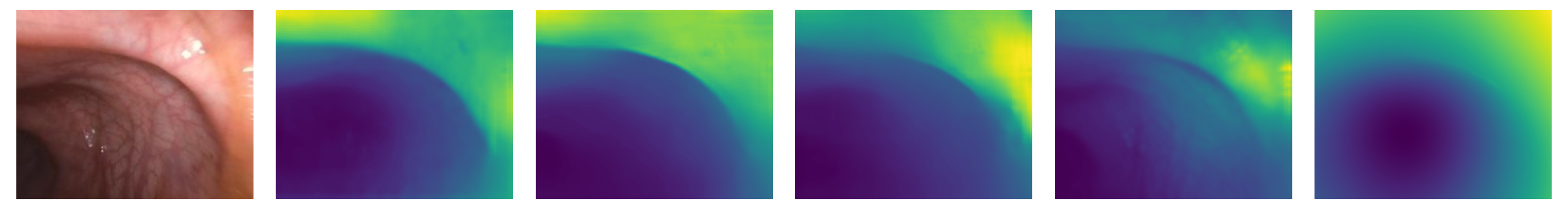}
    \caption{Depth estimation on additional oblique examples.}
    \label{fig:supp_oblique_monodepth}
\end{figure}

\paragraph{Oblique}
In Fig. \ref{fig:depth_demo_monodepth}, we show a few examples of depth estimation using NormDepth and our framework evaluated on images from the proposed oblique test partition (additional examples in Fig. S.\ref{fig:supp_oblique_monodepth}). We have not used masking to prevent depth distortions at specular points. Overall, we see that NormDepth is biased towards predicting a depth depression near the center of the frame and poor predictions near occlusion boundaries. Meanwhile, the baseline model produces significant and repeated errors in the depth map at specular points. Our proposed model produces the best representation of rounded haustral ridges and better distinction between structures. Compared to Ours$_\text{C10K}$ and Ours$_\text{C3VD}$, our model produces depths with stronger discontinuities at occlusion boundaries and overall captures a more nuanced and accurate surface geometry.

\begin{figure}
    \centering
    {\begin{tabularx}{0.8\textwidth}{*6{>{\centering\arraybackslash}X}}
         Input & Ours & Ours$_\text{C10K}$ & Ours$_\text{C3VD}$ & Baseline & NormDepth
    \end{tabularx}}
    \includegraphics[width=0.8\textwidth]{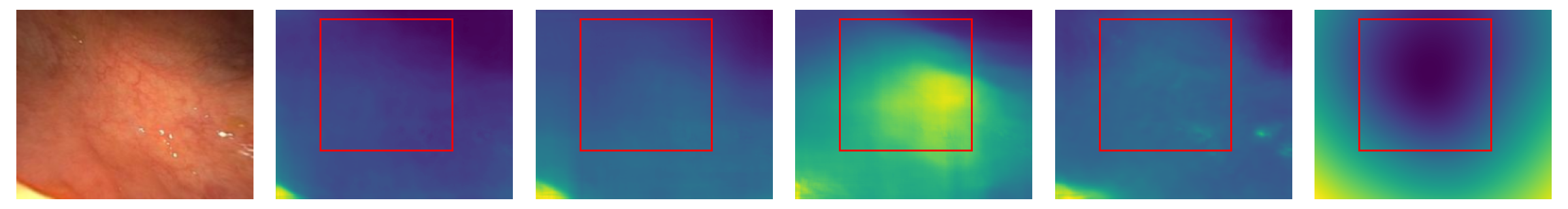}
    \includegraphics[width=0.8\textwidth]{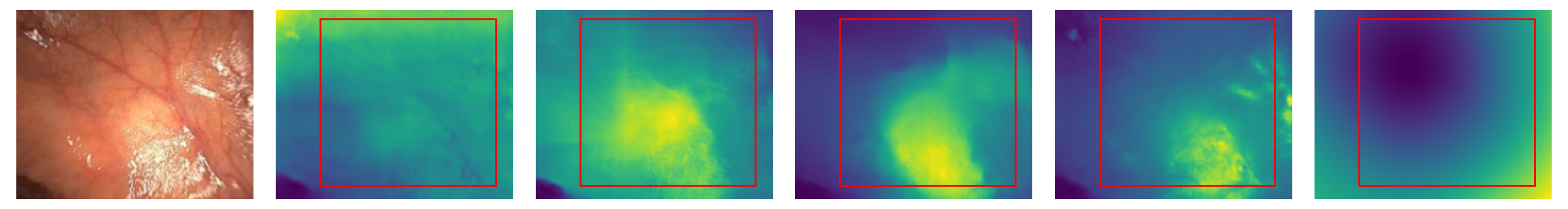}
    \includegraphics[width=0.8\textwidth]{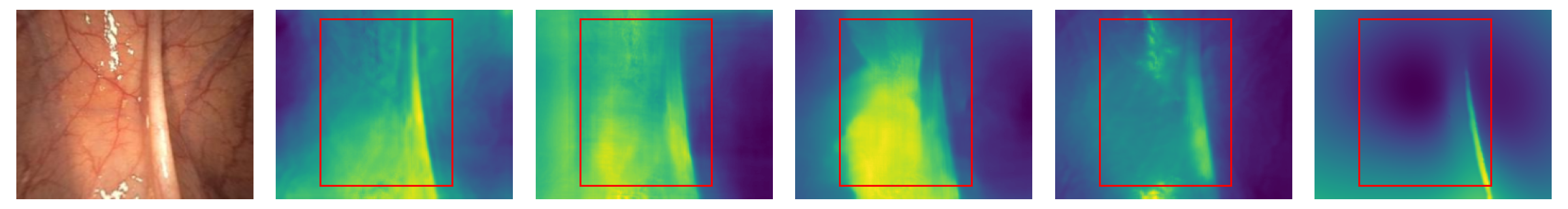}
    \includegraphics[width=0.8\textwidth]{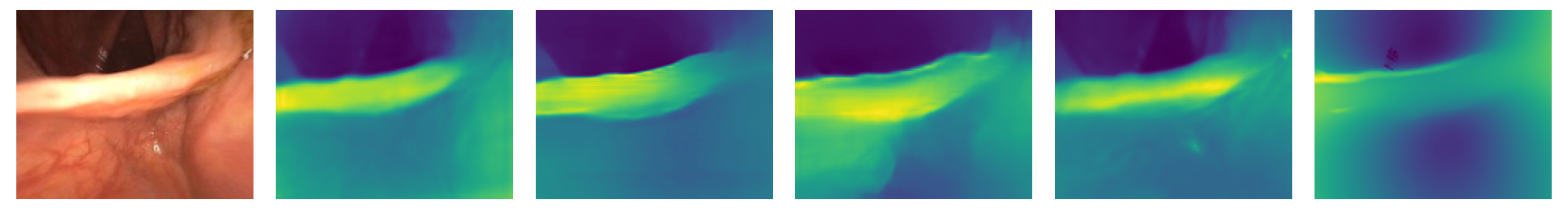}
    \includegraphics[width=0.8\textwidth]{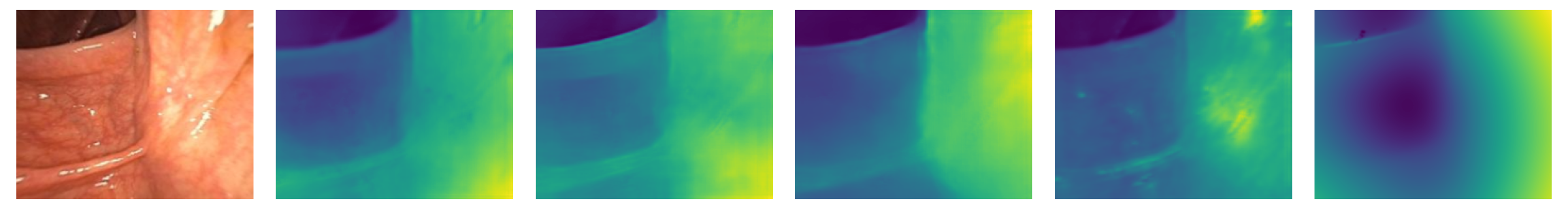}
    \includegraphics[width=0.8\textwidth]{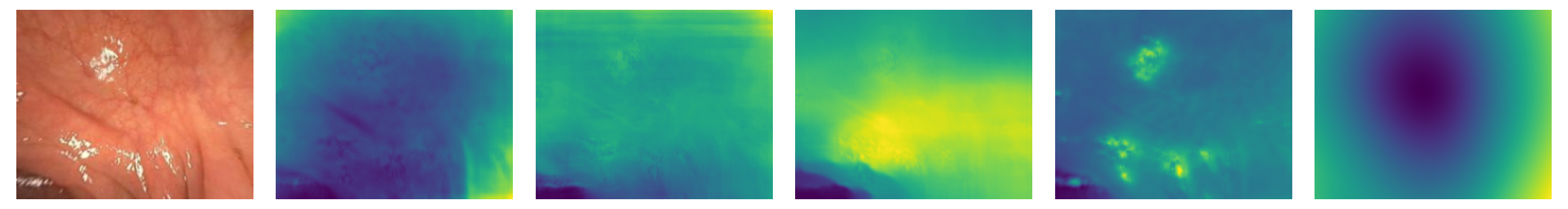}
    \includegraphics[width=0.8\textwidth]{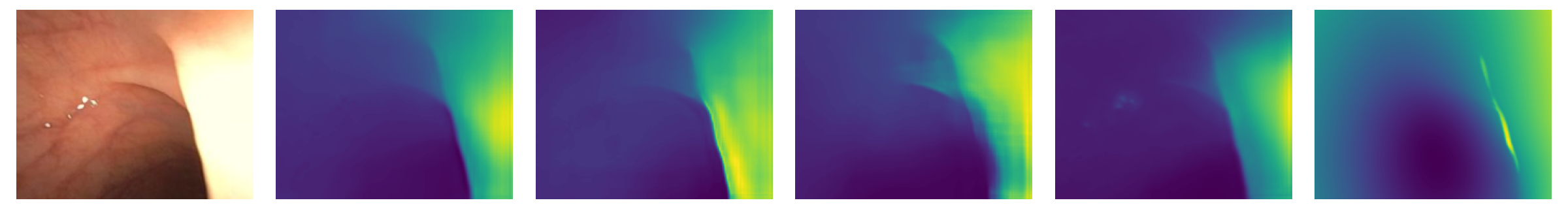}
    \includegraphics[width=0.8\textwidth]{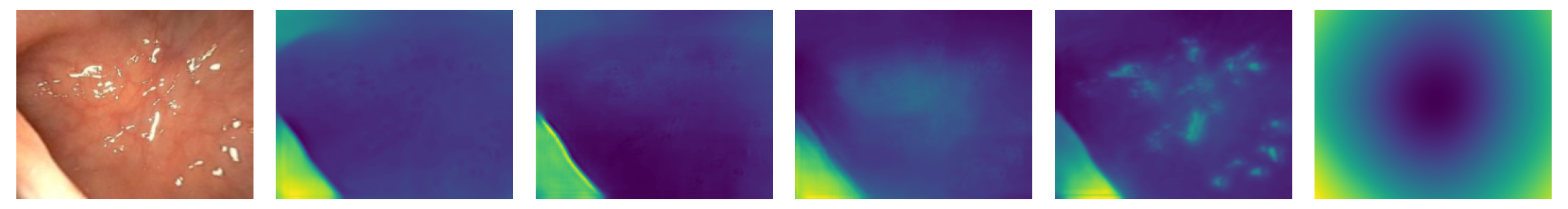}
    \includegraphics[width=0.8\textwidth]{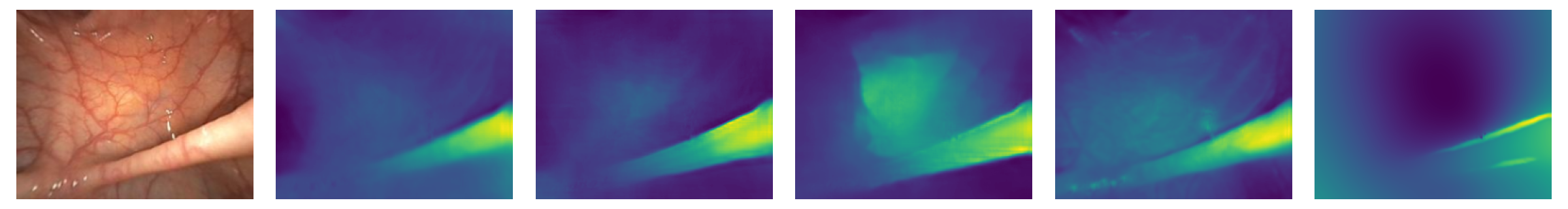}
    \includegraphics[width=0.8\textwidth]{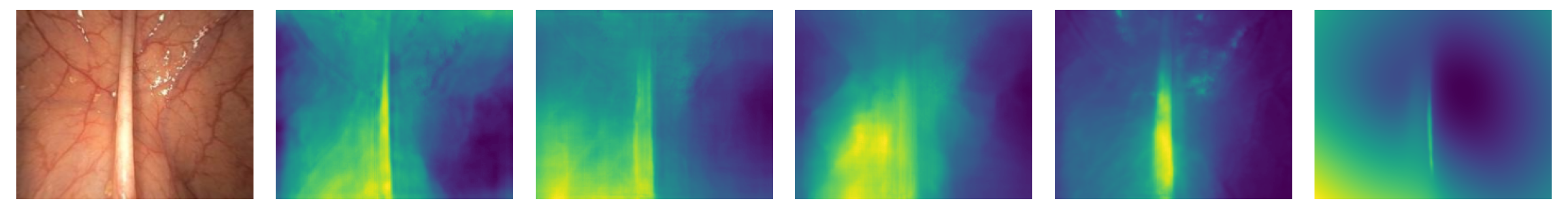}
    \caption{Depth estimation on additional \textit{en face} examples.}
    \label{fig:supp_enface_monodepth}
\end{figure}

\paragraph{En face}
In Fig. \ref{fig:enface_depth_demo_monodepth}, we show a few examples of depth estimation using NormDepth and our framework on images from the proposed \textit{en face} dataset (additional examples in Fig. S.\ref{fig:supp_enface_monodepth}). 
In these examples, the bias of NormDepth towards predicting a center depth depression is particularly evident, as are the failures of the baseline model in specular areas.
Due to the nature of this dataset, there is greater representation of surfaces with strong visual texture from vasculature. Thus we can see that our proposed method has overall improved representation of the overall surface geometry compared to ablations but can also produce distortions to the depth map at regions with strong vascular texture.

\section{Conclusions}
We have demonstrated that structure-preserving sim2real image translation improves monocular depth estimation in challenging colonoscopic frames. To aid this task, we introduce two datasets of hand-picked sequences from clinical data focusing on viewpoints that are under-represented in existing datasets. The image translation results improve texture realism (especially for specular points) while retaining sufficient depth geometry for successful subsequent training of depth estimator networks. We provide evaluation of depth estimation on C3VD and qualitative evaluations on our proposed datasets, finding significant performance improvements on challenging frames using this framework.

\subsection{Limitations and Future Work}
Depth distortions in areas with strongly visible vasculature and few geometric features could be ameliorated by incorporating additional data into the translation target. Future work could focus on applying this approach to pose estimation or other (non-Monodepth2) depth estimation architectures.

\begin{credits}
\subsubsection{\ackname} We thank Stephen Pizer, Sam Ehrenstein, and Julian Rosenman for helpful discussions. Research funding was provided by Olympus Corporation.

\subsubsection{\discintname}
The authors have no competing interests to declare that are relevant to the content of this article.
\end{credits}

\bibliographystyle{splncs04}
\bibliography{main}

\begin{thebibliography}{10}
\providecommand{\url}[1]{\texttt{#1}}
\providecommand{\urlprefix}{URL }
\providecommand{\doi}[1]{https://doi.org/#1}

\bibitem{endomapper}
Azagra, P., Sostres, C., Ferrandez, A., Riazuelo, L., Tomasini, C., Barbed,
  O.L., Morlana, J., Recasens, D., Batlle, V.M., Gómez-Rodríguez, J.J.,
  Elvira, R., López, J., Oriol, C., Civera, J., Tardós, J.D., Murillo, A.C.,
  Lanas, A., Montiel, J.M.M.: Endomapper dataset of complete calibrated
  endoscopy procedures. Scientific Data  \textbf{10}(1) (October 2023).
  \doi{10.1038/s41597-023-02564-7},
  \url{http://dx.doi.org/10.1038/s41597-023-02564-7}

\bibitem{c3vd}
Bobrow, T.L., Golhar, M., Vijayan, R., Akshintala, V.S., Garcia, J.R., Durr,
  N.J.: {Colonoscopy 3D video dataset with paired depth from 2D-3D
  registration}. Medical Image Analysis p. 102956 (2023)

\bibitem{chen-beyond-2022}
Chen, J., Zhang, Z., Xie, X., Li, Y., Xu, T., Ma, K., Zheng, Y.: Beyond mutual
  information: Generative adversarial network for domain adaptation using
  information bottleneck constraint. IEEE Transactions on Medical Imaging
  \textbf{41}(3),  595--607 (2022). \doi{10.1109/TMI.2021.3117996}

\bibitem{cheng_structure-preserving_2020}
Cheng, M.M., Liu, X.C., Wang, J., Lu, S.P., Lai, Y.K., Rosin, P.L.:
  Structure-{Preserving} {Neural} {Style} {Transfer}. IEEE Transactions on
  Image Processing  \textbf{29},  909--920 (2020).
  \doi{10.1109/TIP.2019.2936746},
  \url{https://ieeexplore.ieee.org/document/8816670/}

\bibitem{eftekhar2021omnidata}
Eftekhar, A., Sax, A., Bachmann, R., Malik, J., Zamir, A.: Omnidata: A scalable
  pipeline for making multi-task mid-level vision datasets from 3d scans (2021)

\bibitem{monodepth2}
Godard, C., {Mac Aodha}, O., Firman, M., Brostow, G.J.: Digging into
  self-supervised monocular depth prediction  (October 2019)

\bibitem{colon10k}
Ma, R., McGill, S.K., Wang, R., Rosenman, J., Frahm, J.M., Zhang, Y., Pizer,
  S.: Colon10k: A benchmark for place recognition in colonoscopy. In: 2021 IEEE
  18th International Symposium on Biomedical Imaging (ISBI). pp. 1279--1283
  (2021). \doi{10.1109/ISBI48211.2021.9433780}

\bibitem{mahmood_deep_2018}
Mahmood, F., Durr, N.J.: Deep learning and conditional random fields-based
  depth estimation and topographical reconstruction from conventional
  endoscopy. Medical Image Analysis  \textbf{48},  230--243 (2018).
  \doi{https://doi.org/10.1016/j.media.2018.06.005},
  \url{https://www.sciencedirect.com/science/article/pii/S1361841518303761}

\bibitem{nierengarten-colonoscopy-2023}
Nierengarten, M.B.: Colonoscopy remains the gold standard for screening despite
  recent tarnish. Cancer  \textbf{129} (2023). \doi{10.1002/cncr.34622}

\bibitem{simcol3d}
Rau, A., Bano, S., Jin, Y., Stoyanov, D.: {Simcol3D - 3D Reconstruction during
  Colonoscopy Challenge Dataset}  (September 2023).
  \doi{10.5522/04/24077763.v1}

\bibitem{rau-implicit-2019}
Rau, A., Edwards, P.E., Ahmad, O.F., Riordan, P., Janatka, M., Lova, L.B.,
  Danail, S.: Implicit domain adaptation with conditional generative
  adversarial networks for depth prediction in endoscopy. International Journal
  of Computer Assisted Radiology and Surgery  \textbf{14},  1167--1176 (April
  2019). \doi{10.1007/s11548-019-01962-w}

\bibitem{vanRijn-colonoscopy-2006}
van Rijn, J.C., Reitsma, J.B., Stoker, J., Bossuyt, P.M., van Deventer, S.J.,
  Dekker, E.: Polyp miss rate determined by tandem colonoscopy: a systematic
  review. The American Journal of Gastroenterology  (2006).
  \doi{10.1111/j.1572-0241.2006.00390.x}

\bibitem{siegel-cancer-2024}
Siegel, R.L., Giaquinto, A.N., Jemal, A.: Cancer statistics. CA: A Cancer
  Journal for Clinicians  \textbf{74} (2024). \doi{10.3322/caac.21820}

\bibitem{vemulapalli-polyps-2022}
Vemulapalli, K.C., Lahr, R.E., Rex, D.K.: Most large colorectal polyps missed
  by gastroenterology fellows at colonoscopy are sessile serrated lesions.
  Endoscopy International Open  (2022). \doi{10.1055/a-1784-0959}

\bibitem{wang-surface-2023}
Wang, S., Zhang, Y., McGill, S.K., Rosenman, J.G., Frahm, J.M., Sengupta, S.,
  Pizer, S.M.: A surface-normal based neural framework for colonoscopy
  reconstruction. In: Information Processing in Medical Imaging (2023)

\bibitem{depthanything}
Yang, L., Kang, B., Huang, Z., Xu, X., Feng, J., Zhao, H.: Depth anything:
  Unleashing the power of large-scale unlabeled data. arXiv:2401.10891  (2024)

\bibitem{yoon_colonoscopic_2022}
Yoon, D., Kong, H.J., Kim, B.S., Cho, W.S., Lee, J.C., Cho, M., Lim, M.H.,
  Yang, S.Y., Lim, S.H., Lee, J., Song, J.H., Chung, G.E., Choi, J.M., Kang,
  H.Y., Bae, J.H., Kim, S.: Colonoscopic image synthesis with generative
  adversarial network for enhanced detection of sessile serrated lesions using
  convolutional neural network. Scientific Reports  \textbf{12}(1), ~261
  (January 2022). \doi{10.1038/s41598-021-04247-y},
  \url{https://www.nature.com/articles/s41598-021-04247-y}, number: 1
  Publisher: Nature Publishing Group

\bibitem{zhang-sydneysimulator-2021}
Zhang, S., Zhao, L., Huang, S., Ye, M., Hao, Q.: A template-based 3d
  reconstruction of colon structures and textures from stereo colonoscopic
  images. IEEE Transactions on Medical Robotics and Bionics  \textbf{3}(1),
  85--95 (2021). \doi{10.1109/TMRB.2020.3044108}

\bibitem{CycleGAN2017}
Zhu, J.Y., Park, T., Isola, P., Efros, A.A.: Unpaired image-to-image
  translation using cycle-consistent adversarial networkss. In: Computer Vision
  (ICCV), 2017 IEEE International Conference on (2017)

\end{thebibliography}

\newpage

\section*{Supplementary Material}

\end{document}